\documentclass{osa-article}
\journal{osac}
\articletype{Research Article}

\usepackage{siunitx}

\AtBeginDocument{\usepackage{booktabs}}  % Necessary to work with Revtex class

\newcommand{\blue}[1]{{\color{blue} #1}}

\usepackage{layouts}

\begin{document}

\title{Silicon Photonic Architecture for Training Deep Neural Networks with Direct Feedback Alignment}

\author{
Matthew J. Filipovich,\authormark{1,*, $\dag$} 
Zhimu Guo,\authormark{1, $\dag$} 
Mohammed Al-Qadasi,\authormark{2} 
Bicky A. Marquez,\authormark{1} 
Hugh D. Morison,\authormark{1} 
Volker J. Sorger,\authormark{3} 
Paul R. Prucnal,\authormark{4} 
Sudip Shekhar,\authormark{2} 
and Bhavin J. Shastri\authormark{1,5,*} 
}

\address{
\authormark{1}Department of Physics, Engineering Physics and Astronomy, Queen's University, Kingston, ON K7L 3N6, Canada\\
\authormark{2}Department of Electrical and Computer Engineering, University of British Columbia, Vancouver, BC V6T 1Z4, Canada\\
\authormark{3}Department of Electrical and Computer Engineering, George Washington University, Washington, DC 20052, USA\\
\authormark{4}Department of Electrical Engineering, Princeton University, Princeton, NJ 08542, USA\\
\authormark{5}Vector Institute, Toronto, ON M5G 1M1, Canada\\
\authormark{$\dag$} These authors contributed equally to this paper.

}

\email{\authormark{*}matthew.filipovich@queensu.ca, shastri@ieee.org} %% email address is required



\begin{abstract*}
There has been growing interest in using photonic processors for performing neural network inference operations; however, these networks are currently trained using standard digital electronics. Here, we propose on-chip training of neural networks enabled by a CMOS-compatible silicon photonic architecture to harness the potential for massively parallel, efficient, and fast data operations. Our scheme employs the direct feedback alignment training algorithm, which trains neural networks using error feedback rather than error backpropagation, and can operate at speeds of trillions of multiply-accumulate (MAC) operations per second while consuming less than one picojoule per MAC operation. The photonic architecture exploits parallelized matrix-vector multiplications using arrays of microring resonators for processing multi-channel analog signals along single waveguide buses to calculate the gradient vector for each neural network layer in situ. We also experimentally demonstrate training deep neural networks with the MNIST dataset using on-chip MAC operation results. Our novel approach for efficient, ultra-fast neural network training showcases photonics as a promising platform for executing AI applications. 
\end{abstract*}

\section{Introduction}
Propelled by recent advances in deep learning \cite{Deep_Learning_2015}, the fields of artificial intelligence (AI) and neuromorphic computing (neuro-biological computer architectures) have seen a renaissance over the past decade \cite{Schuman_2017}. Today, the software implementations of AI algorithms are executed on traditional computers based on the von Neumann architecture \cite{von1993first}. However, this architecture faces inherent data-transfer speed limitations due to the separation of memory and processor unit, known as the von Neumann bottleneck. Neuromorphic computing seeks to eliminate this constraint by exploiting the underlying elementary physics of hardware systems to create an isomorphism with AI algorithms \cite{Mead_1990}. As Moore’s law (which claims that the number of transistors on a microchip doubles every two years) approaches saturation and with the exponentially increasing needs of advanced machine learning algorithms \cite{mehonic_brain-inspired_2022, Canziani_2017}, neuromorphic computing has attracted renewed interest as an alternative to the standard electronic digital computer architectures \cite{Shastri_nature}.

Silicon photonics has shown to be a promising platform for developing neuromorphic systems due to its compatibility with the mature silicon integrated circuit industry and the availability of high-quality silicon-on-insulator (SOI) wafers \cite{Prucnal_2017, Chrostowski_2015}. Monolithic silicon photonics enables the integration of active photonic and electronic components onto a single photonic integrated circuit (PIC), including modulators, detectors, amplifiers, complementary metal-oxide-semiconductor (CMOS) control circuits, and optical multiplexers \cite{Bogaerts_Chrostowski_2018, thomson_roadmap_2016}. Compared to their electronic counterparts for neuromorphic applications, photonic systems offer low-latency, high-bandwidth density and baud rates, low-cost communication, and inherent parallelism using optical multiplexing \cite{Lima_2019, Peng_Nahmias_2018}.

\begin{figure}
    \centering
    \includegraphics[width=0.95\linewidth]{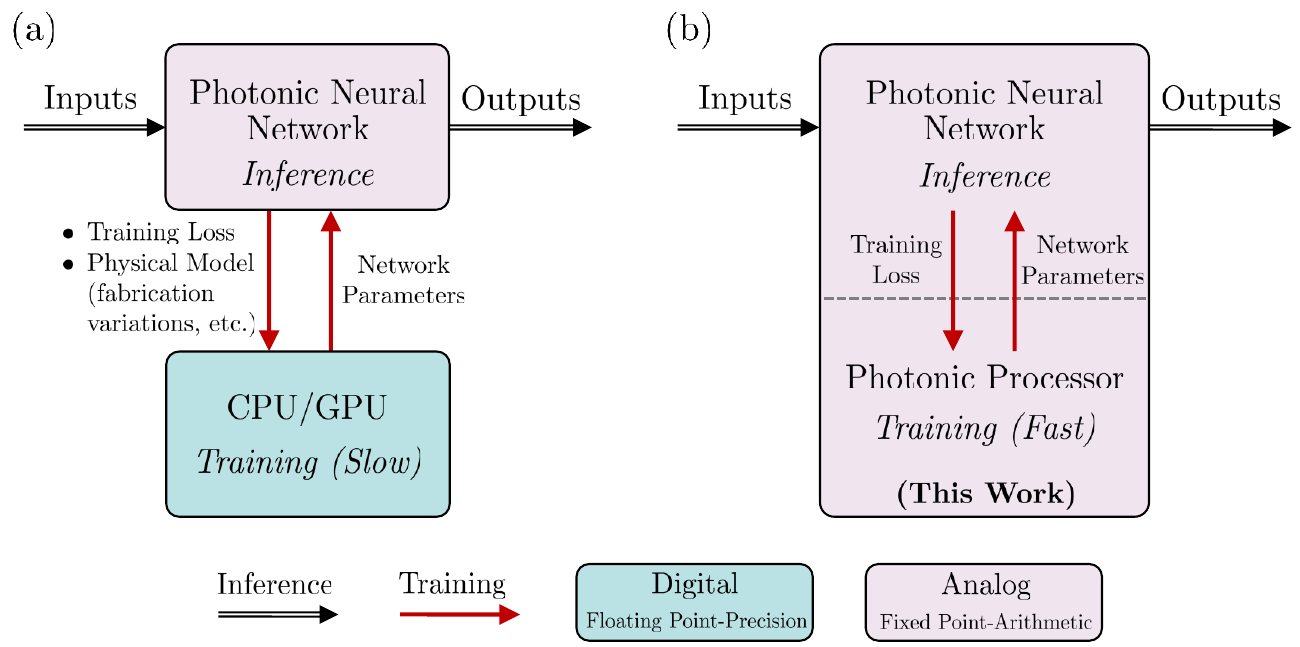}
    \caption{Training photonic neural networks offline (slow, floating-point precision) and in situ (fast, fixed-point arithmetic). (a) Offline training using a digital platform (CPU/GPU) to model the analog hardware system and determine the optimized parameters, which are then mapped to the hardware-based neural network for inference. (b) In situ training using photonic hardware to directly determine the optimized analog neural network parameters \cite{hughes_training_2018, zhou_situ_2020, guo_backpropagation_2021, pai_experimentally_2022, bandyopadhyay_single_2022}.}
    \label{fig:offline_online_training}
\end{figure}

\begin{figure}
    \centering
    \includegraphics[width=0.95\linewidth]{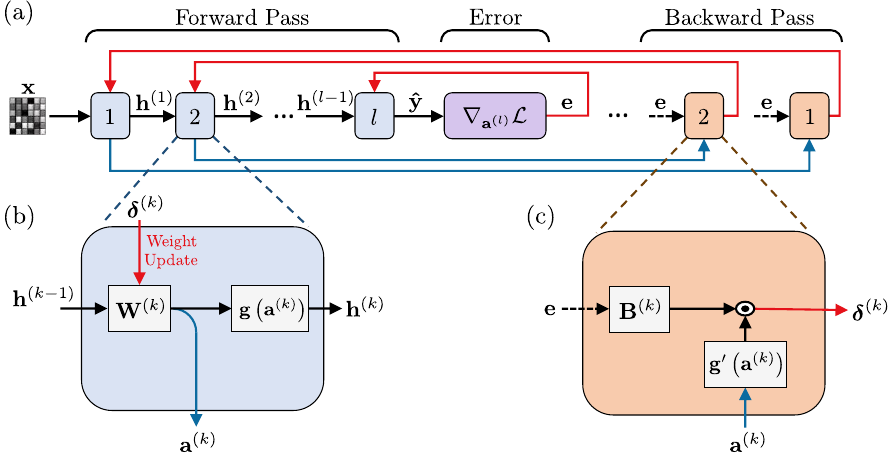}
    \caption{Schematic illustration of the DFA algorithm. (a)~The input vector $\mathbf{x}$ propagates through the neural network with $l$~layers during the forward pass (inference), where $\mathbf{h}^{(k)}$ is the output from the hidden layer $k$ and $\mathbf{\hat{y}}$ is the output vector. The error vector $\mathbf{e}$ is then calculated, defined as the gradient of the loss function $\mathcal{L}\left(\mathbf{\hat y}, \mathbf{y}\right)$ where $\mathbf{y}$ is the target. During the backward pass, the error is propagated through fixed random feedback connections directly from the output layer to the hidden layers. The gradient vector $\boldsymbol{\delta}^{(k)}$ for each hidden layer $k$ is calculated and used for the weight update (the output layer weight matrix $\mathbf{W}^{(l)}$ is updated using the error vector $\mathbf{e}$). (b)~Block diagram of the operations performed during the forward pass at the layer $k$. (c)~Block diagram of the gradient vector calculations performed during the backward pass at the hidden layer $k$.}
    \label{fig:dfa}
\end{figure}

One area of machine learning that would benefit from the low-power consumption and high-information processing bandwidth enabled by photonics is the training of large neural networks. Several photonic architectures have been proposed for executing in-memory computation of neural network inference \cite{Tait_2017, Shen_2017, Tait_2019, Zuo_2019, spall_fully_2020}. However, for neural networks to perform useful tasks, the optimal network parameters (weights and biases) must first be determined using deep learning training algorithms. These algorithms have high-computation and memory costs that pose challenges to the current hardware platforms executing them \cite{Esser_2016}, and the substantial energy consumption required to train large neural networks using standard von Neumann architectures presents a significant financial and environmental cost \cite{Strubell_Ganesh_McCallum_2019}. Currently, analog hardware-based neural networks are primarily trained offline (see Fig.~\ref{fig:offline_online_training}(a))  using training algorithms implemented on digital platforms (optimized for executing sequential, procedure-based programs) which model the analog system and then map the optimized network parameters to the hardware platform for inference. However, the models may not capture all the manufacturing imperfections and variations of the physical neurons and their interconnections, including the inherent noise in the system. Additionally, the computational overhead required to accurately model the physical system can result in slow execution times compared to \textit{in situ} training (see Fig.~\ref{fig:offline_online_training}(b)).

The recently proposed direct feedback alignment (DFA) \cite{nokland2016direct} supervised learning algorithm has gathered interest as a bio-plausible alternative to the popular backpropagation training algorithm \cite{Rumelhart_1986}. The DFA algorithm, shown in Fig.~\ref{fig:dfa} and further described in Supplement~1, is a supervised learning algorithm that propagates the error through fixed random feedback connections directly from the output layer to the hidden layers during the backward pass \cite{nokland2016direct}. Unlike backpropagation, the DFA algorithm does not require network layers to be updated sequentially during the backward pass, enabling the algorithm to be a suitable candidate for efficient parallelization using photonics. The algorithm has been used to train neural networks using the MNIST, CIFAR-10, and CIFAR-100 datasets, and yields comparable performance to backpropagation \cite{nokland2016direct}. As well, the DFA algorithm has been shown to obtain performances comparable to fine-tuned backpropagation in applications requiring state-of-the-art deep learning networks, including click-through rate prediction with recommender systems and neural view synthesis with neural radiance fields \cite{DFA_scaling}. A recent theory suggests that training shallow networks with the DFA algorithm occurs in two steps: the first step is an alignment phase where the weights are modified to align the approximate gradient with the true gradient of the loss function, which is followed by a memorization phase where the model focuses on fitting the data \cite{Refinetti_2020}.

In this work, we 1) propose a silicon photonic architecture that implements a training algorithm (DFA) on-chip; 2) show that DFA is well suited for photonic implementation because all the network layers can be updated in parallel during the backward pass (as the same error is propagated to each layer); and 3) demonstrate that in situ training with DFA is robust to noise as hardware nonidealities are inherently accounted for. Our proposed silicon photonic architecture uses an electro-optic circuit for calculating the gradient vector for each hidden layer in situ, which is the most computationally expensive operation performed during the backward pass. The proposed architecture exploits the speed and energy advantages of photonics to determine the gradient vector for each hidden layer in a single operational cycle. The scope of this paper is limited to the implementation of the DFA algorithm's backward pass; however, inference can also be performed using a similar photonic architecture \cite{Tait_2017}. During the backward pass, the error from the network's inference step is encoded on multi-channel optical inputs. The electro-optic circuit then calculates the gradient vector for each hidden layer, which is used to update the network parameters stored in memory using an external digital control system. We also demonstrate training feed-forward neural networks on the MNIST dataset using experimental on-chip results to validate the photonic architecture.

\section{Multiply-accumulate operations in photonics}\label{sec:MAC}
\begin{figure}
    \centering
    \includegraphics[width=\linewidth]{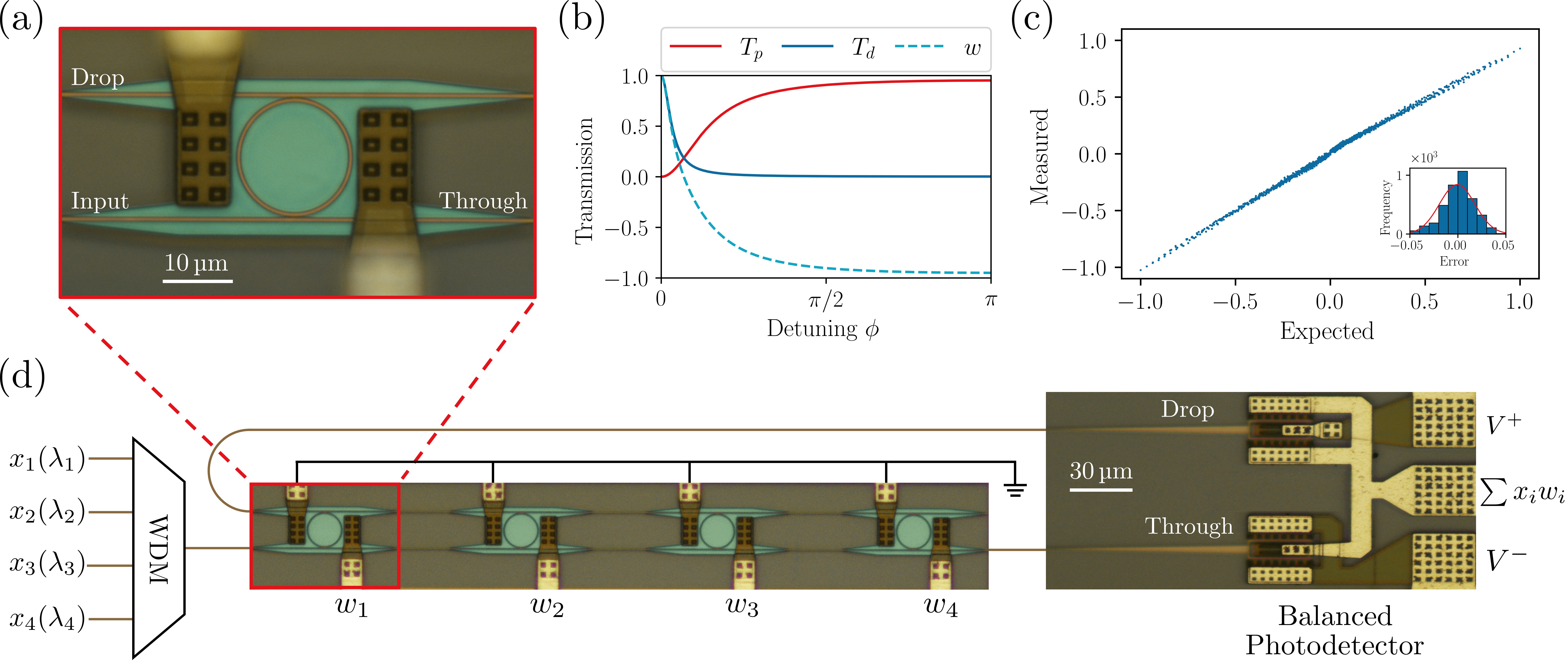}
    \caption{MAC operations performed using integrated photonic devices. (a)~MRR in add-drop configuration with $\SI{8}{\micro\metre}$ radius coupled to through and drop ports. (b)~Theoretical transmission profile of an add-drop MRR with a self-coupling coefficient of 0.95 and negligible attenuation. The weighting value $w$ is given by $T_d-T_p$. (c)~Experimental multiplication operations performed using a single MRR across 3900 random combinations of the input and weight values. The error in the  multiplication results has a standard deviation of 0.019 (effective resolution of 6.72 bits) and mean of $-0.001$. (d)~$1\times4$ MRR array where the coupled through and drop ports are connected to a BPD. The inner product operation is executed in the analog domain by injecting multi-channel analog signals with wavelength $\lambda_i$ and amplitude $x_i$. Each MRR in the array is tuned to weight a corresponding wavelength by $w_i$, which yields the inner product $\mathbf{x}\cdot\mathbf{w}$ in the electrical domain.}
    \label{fig:mrrs}
\end{figure}

The proposed architecture uses microring resonators (MRRs), which are closed-loop waveguides that use codirectional evanescent coupling between the ring and adjacent bus waveguides, as tunable filters to weight photonic signals through amplitude modulation \cite{Bogaerts_2012}. Modulating the transmission of optical signals for specified wavelengths using silicon-based MRRs is achieved by changing their refractive index which tunes the resonance peaks. The refractive index can be changed by either varying the concentration of carriers through external biasing (e.g., carrier depletion in a PN-junction, carrier injection in a PIN junction) or by exploiting the thermo-optic effect~\cite{Reed_Mashanovich_Gardes_Thomson_2010}. In this paper, we experimentally demonstrate thermally tuned MRRs using in-ring N-doped photoconductive heaters; however, we also discuss the implementation of efficient, high-speed MRRs tuned using carrier depletion. In our proposed architecture, the MRRs that perform multiply-accumulate (MAC) operations are arranged in add-drop geometries coupled to two waveguide buses, known as the through and drop ports, as shown in Fig.~\ref{fig:mrrs}(a). Tuning the MRR's resonance to control the amount of light transmission through the two ports effectively weights the input optical signal.

The inner product operation, which consists of a series of MAC operations $\mathbf{x} \cdot \mathbf{w} = \Sigma x_i w_i$, is executed in photonics using an array of MRRs and a balanced photodetector (BPD) \cite{Tait_MRR_2016,Tait_2017}. Using wavelength-division multiplexing (WDM) techniques to process multi-channel analog signals along the same waveguide bus, where each optical signal with wavelength $\lambda_i$ is amplitude encoded with a value $x_i$, each MRR in the array is tuned with a specified weighting value $w_i\in[-1,1]$ for the corresponding optical input wavelength $\lambda_i$. The through and drop port transmission spectrums, $T_p$ and $T_d$, of each MRR as a function of round-trip phase shift are Lorentzian-shaped and centered at the resonance phase with the incoming light. When the coupling losses between the MRR and bus waveguides are negligible, the relationship between the through and drop port transmissions for each MRR is $T_p=1-T_d$. These two ports are connected into a BPD which implements an electro-optic transfer function proportional to $|E_0|^2(T_d-T_p)$, where $E_0$ is the amplitude of the input optical signal. Thus, the assigned weighting value $w_i$ of each MRR is given by $T_d-T_p$. Using this scheme, external biasing of the MRR tunes the resonance peak to correspond with the desired weighting for a selected optical wavelength, as shown in Fig.~\ref{fig:mrrs}(b). Finally, the inner product between two vectors of size $N$ can be executed using an array of $N$ MRRs that are coupled to the same through and drop ports, as shown in Fig.~\ref{fig:mrrs}(d). 

Experimental multiplication operation results using a single MRR are shown in Fig.~\ref{fig:mrrs}(c). The experimental data were collected by varying the encoded input signal and MRR weighting values across random combinations while measuring the optical power in the drop and through ports ($P_d$ and $P_t$) using a power meter. The output optical power for each combination was measured three separate times and the average value was recorded. The multiplication results from the two operands were determined by calculating $P_d-P_t$ off-chip, and the results were scaled to match the expected output range of values between $-$1 and 1.

The continuous and multi-channel control of MRR arrays based on feedback control is an ongoing area of research, and a record-high accuracy of 9 bits for a single MRR using thermal tuning has recently been observed with negligible inter-channel crosstalk \cite{zhang_silicon_2022}. Due to precision limitations in PIC manufacturing leading to device-to-device variations, the relationship between the applied MRR bias and the change in weighting value for a specified optical wavelength must be determined experimentally  \cite{TaitFeedback_2018, Huang_2020}. The current tuning approach relies on feedforward control to calibrate the MRRs, as well as feedback control for sensing the state of the system and correcting for any changes due to dynamic variability, such as ambient environmental fluctuations \cite{Jayatilleka_2019}.

\section{Photonic deep learning architecture}\label{architecture}

\begin{figure}[t]
    \centering
    \includegraphics[width=\linewidth]{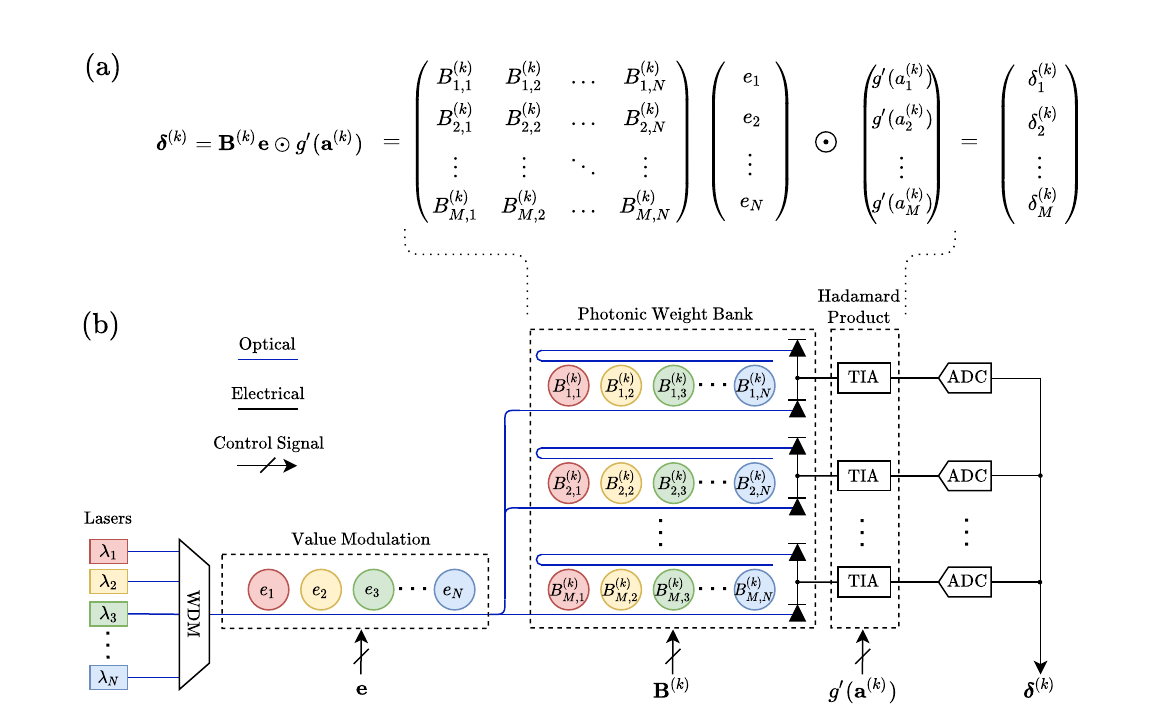}
    \caption{Schematic of our proposed photonic DFA architecture. (a)~Equation defining the gradient vector $\boldsymbol{\delta}^{(k)}$ for the hidden layer $k$. (b)~Silicon photonic architecture for calculating the gradient vector $\boldsymbol{\delta}^{(k)}$ for the hidden layer $k$. The output layer gradient vector $\mathbf{e}$ is amplitude encoded onto $N$ different wavelengths using an array of MRRs, and the matrix-vector product of the $M\times N$ matrix $\mathbf{B}^{(k)}$ and vector $\mathbf{e}$ is performed in the photonic weight bank. The Hadamard product between the vectors $\mathbf{B}^{(k)}\mathbf{e}$ and $g'(\mathbf{a}^{(k)})$ is executed using TIAs in the electrical domain, yielding the desired vector $\boldsymbol{\delta}^{(k)}$ which is converted to the digital domain using ADCs. At each time step, the control signals that encode the MRRs and TIAs are fetched from memory.}
    \label{fig:photonic_architecture}
\end{figure}

The proposed architecture for executing the DFA training algorithm exploits MAC operations performed in photonics (as discussed in the previous section) to execute matrix-vector multiplications in the analog domain during the backward pass, which are the most computationally expensive operation performed during training. For each training example during the backward pass, the architecture calculates the gradient vector $\boldsymbol{\delta}^{(k)}$ for the hidden layer $k$ in a single operational cycle. The gradient vector (see Fig.~\ref{fig:photonic_architecture}(a)) is given by
\begin{equation}
    \boldsymbol{\delta}^{(k)} = \mathbf{B}^{(k)} \mathbf{e} \odot g'\left(\mathbf{a}^{(k)}\right),
\end{equation}
where $\mathbf{B}^{(k)}$ is a fixed random weight matrix with appropriate dimensions for the hidden layer $k$, $\mathbf{e}$ is the error from the gradient of the loss function, $\odot$ is the Hadamard product (element-wise multiplication operator), and $g'$ is the derivative of the activation function with respect to $\mathbf{a}^{(k)}$, which is the sum of the weighted input signals in the hidden layer $k$. The network's weights and biases are then updated using the calculated gradient vector $\boldsymbol{\delta}^{(k)}$ for each hidden layer $k$. The cost of updating the network's parameters can be amortized using mini-batches during training.

A schematic of the proposed silicon photonic DFA architecture, which can be implemented on a PIC, is shown in Fig.~\ref{fig:photonic_architecture}(b). The DFA architecture requires a digital control system to tune the active electro-optic components on-chip, which include the MRRs that modulate the incoming laser light with the error vector $\mathbf{e}$ and the transimpedance amplifiers (TIAs) with tunable gain to convert photocurrent to voltage and scale it to implement the Hadamard product. To execute the DFA algorithm, a training example is first propagated through a neural network of variable size. This inference operation can be performed using an analog photonic circuit  \cite{Tait_2017} or a separate digital processor. The error vector $\mathbf{e}$ from the derivative of the loss function is then calculated using a dedicated CMOS processor that can be integrated on-chip. Finally, the electro-optic circuit calculates the DFA gradient vector~$\boldsymbol{\delta}^{(k)}$ for each hidden layer $k$, which is used to update the network parameters. 

To calculate the gradient vector $\boldsymbol{\delta}^{(k)}$ using the electro-optic circuit, WDM is used to combine $N$ laser signals with different wavelengths onto a single waveguide. The element values in the error vector $\mathbf{e}$ are amplitude encoded onto the $N$ optical signals using an array of $N$ all-pass MRRs, where each MRR is tuned for a corresponding optical wavelength. The intensities of the input optical signals are identical to allow an encoding scheme that linearly maps the amplitude modulation value to the through port transmission. The modulated optical signals representing the error vector $\mathbf{e}$ are then coupled into arrays of parallel MRRs, herein referred to as the photonic weight bank, which perform the required matrix-vector product with the matrix  $\mathbf{B}^{(k)}$  \cite{Tait_2019}. The photonic weight bank consists of $M$ rows of MRR arrays with $N$ MRRs per row, where the incoming signals are coupled evenly into each row using 1$\times M$ symmetrical optical splitters \cite{Sridarshini}. As discussed in the previous section for performing MAC operations in photonics, each MRR in the photonic weight bank is arranged in an add-drop configuration coupled to two separate waveguide buses connected to a BPD. To perform the matrix-vector product, each column of MRRs in the photonic weight bank is tuned for the corresponding optical wavelength $\lambda_n$, and thus the element $B^{(k)}_{m,n}$ is mapped to the MRR in the $m$th row and $n$th column. Using the add-drop configuration, the incoming optical signals, which correspond to the error vector $\mathbf{e}$, are weighted by modifying the optical transmissions, $T_p$ and $T_d$, for the through and drop ports of the MRRs. Summing the two transmission ports in the electrical domain allows the MRRs to be encoded with a weighting $w\in[-1,1]$, assuming there is negligible loss in the system \cite{Bangari_2020}. If the size of the photonic weight bank is larger than the dimensions of the matrix $\mathbf{B}^{(k)}$, the redundant MRRs can be tuned with a weighting of zero. A negative value in the error vector can be encoded in the architecture by inverting the sign of the inscribed weighting values of the corresponding column of MRRs in the photonic weight bank.

Finally, the Hadamard product between the two vectors $\mathbf{B}^{(k)}\mathbf{e}$ and $g'(\mathbf{a}^{(k)})$ is performed using a set of TIAs, where each BPD output is connected to a TIA. The vector $g'(\mathbf{a}^{(k)})$ is calculated by the dedicated monolithic CMOS processor (the vector $\mathbf{a}^{(k)}$ is known from the inference step) and encoded onto the voltage signals, supplied by the digital control system, that set the gain of the TIAs. The elements in the vector $g'(\mathbf{a}^{(k)})$ are binary (0 or 1) when the ReLU function is used for the activation function $g(\cdot)$. Since the vector $\mathbf{a}^{(k)}$ was previously calculated during the forward inference process, setting the gain does not impede the system's maximum operating speed. The analog electrical signals are then converted to digital values with analog-to-digital converters (ADCs) and are used by the digital control system to update the neural network's parameters. 

The size of the photonic weight bank is physically bounded by the dimensions of the PIC and the maximum number of supported WDM channels in a single waveguide. The WDM channel limit is dependent on the finesse of the MRRs and the channel spacing, and an optimized design of the MRRs with a finesse of 368 could support up to 108 distinct channels \cite{Tait_MRR_2016}. However, the dimensions of the photonic weight bank do not restrict the size of the neural network being trained: a customized general matrix multiplication (GeMM) compiler can be used to subdivide the matrix $\mathbf{B}^{(k)}$
%if it is larger than the dimensions of the photonic weight bank,  
such that the matrix-vector product is determined over multiple operational cycles by calculating a subset of the output vector at each cycle~\cite{guo_multi-level_2022, ma_high-density_2022}. Indeed, the compatibility of GeMM compilers with the photonic DFA circuit could enable on-chip training of large state-of-the-art neural network architectures by reducing the required mathematical operations at each cycle based on the topology of the PIC.

%The field of silicon photonics has rapidly accelerated over the last several years; however, many considerable challenges must be addressed before the widespread adoption of optical technologies for AI applications. 
The co-integration of active on-chip silicon electronics with silicon photonics is crucial for the calibration and control of the photonic components in the proposed architecture~\cite{Bogaerts_Chrostowski_2018}. Monolithic fabrication processes, which integrate electronics and photonics on the same substrate, are being investigated as the demand for silicon photonics technology grows~\cite{Giewont_2019}. An alternative near-term implementation of the photonic architecture could use flip-chip bonding by fabricating two dies (one optimized for silicon photonics and the other for CMOS electronics) and soldering them together. Another challenge facing silicon photonic computational architectures is their current reliance on off-chip light sources connected through fiber packaging. Several recent approaches have been proposed to integrate light sources directly onto silicon waveguide layers, including strain engineering of germanium and all-silicon emissive defects and rare-earth-element doping \cite{Buckley_2017, Zhou_2015}. An alternative approach, which is well suited for our highly-parallelized photonic architecture and has recently been experimentally demonstrated for performing photonic in-memory computing~\cite{Feldmann_2021}, is the use of an on-chip frequency comb source that generates evenly spaced emission wavelengths~\cite{Lipson_2019}. However, the optical source must be highly efficient while producing enough power for the entire system, which includes the requirement to overcome the capacitance and shot noise of the photodetectors.

\section{DFA training experiment}
\begin{figure}
    \centering
    \includegraphics[width=0.6\linewidth]{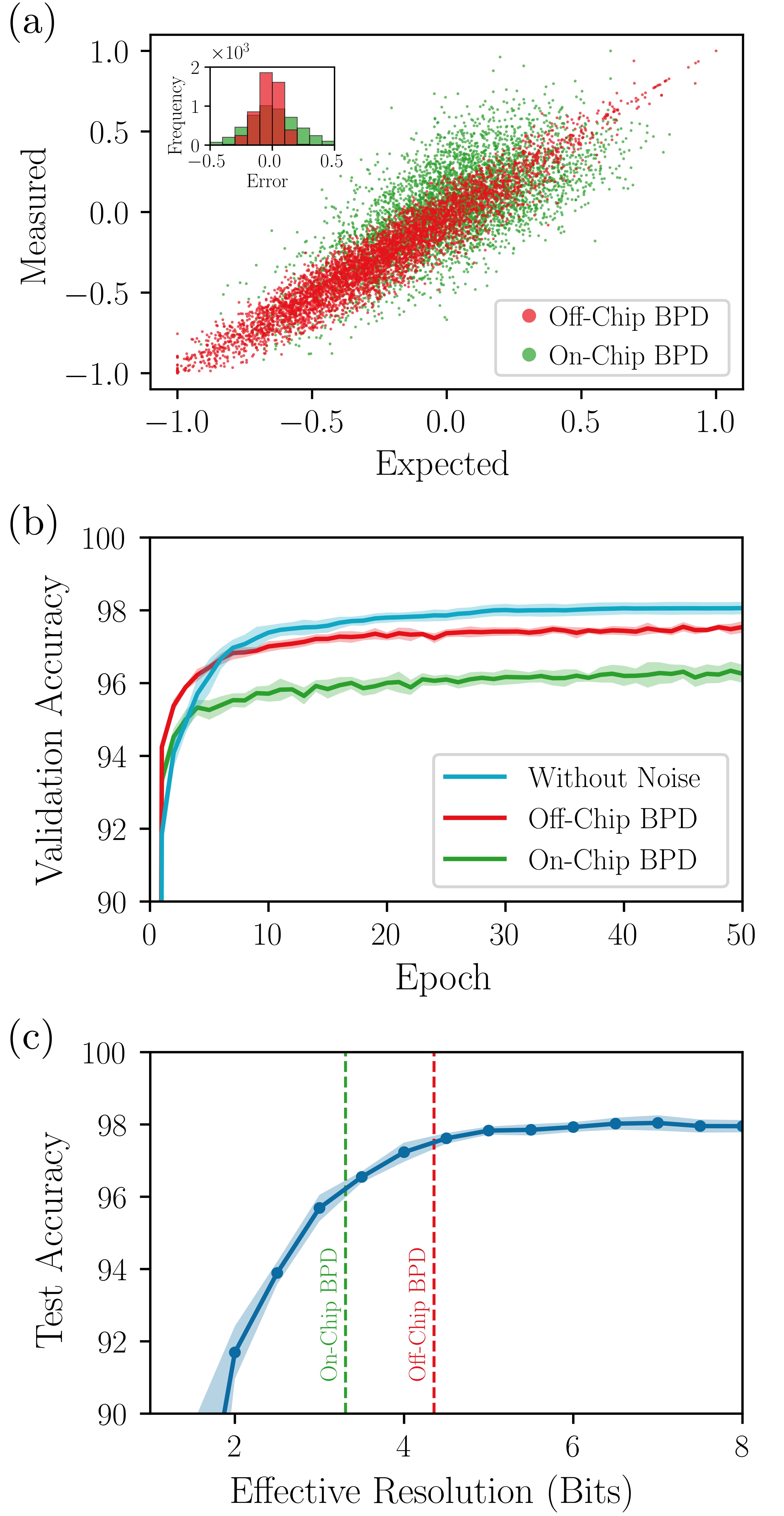}
    \caption{
    Experimental demonstration of neural network training on the MNIST dataset using two photonic circuits with different characteristic noise. 
    (a)~Two separate sets of 5000 photonic inner product measurements using $1\times4$ MRR arrays connected to both off-chip and on-chip BPD circuits. 
    The error, defined as the measured minus expected values, from the off-chip BPD (on-chip BPD) has a standard deviation of 0.098 (0.202) and mean of 0.003 (0.003), yielding an effective resolution of 4.35 bits (3.31 bits). 
    The accuracy of the on-chip BPD circuit measurements could be improved using a correctly configured BPD control circuit, as demonstrated using the off-chip BPD from Thorlabs.
    (b)~Validation accuracy during training on the MNIST dataset using experimental inner product operation results, as well as without noise.
    (c)~Test accuracy on the MNIST dataset as a function of the effective resolution used in the gradient calculations. The dashed red and green lines are the effective resolutions achieved by the off-chip BPD circuit (test accuracy of $97.41 \pm 0.15\%$) and on-chip BPD circuit ($96.33\pm 0.16\%$), respectively. 
    }
    \label{fig:experimental}
\end{figure}

We performed two sets of photonic inner product operation measurements using two similar circuits (with inherently different noise) to validate our proposed architecture. We then used these results to train feed-forward neural networks on the MNIST dataset in a Python simulation. Both circuits contain identical $1\times4$ MRR arrays with coupled through and drop ports; the ports in one circuit connect to an on-chip BPD (shown in Fig.~\ref{fig:mrrs}(d)), while the ports in the other circuit connect to grating couplers for off-chip photodetection using a 5~GHz BPD from Thorlabs (part BDX1BA). The MRRs are thermally tuned using in-ring N-doped photoconductive heaters \cite{Jayatilleka_2015, Jayatilleka_2019} and their add-drop configuration enables positive and negative weighting value encoding, as described in Sec.~\ref{sec:MAC}.  We used four external cavity laser sources with different wavelengths (1546.558~nm, 1548.675~nm, 1549.595~nm, and 1551.480~nm)  modulated directly by the laser source using embedded electronic modulation, and the different optical power levels were mapped to the expected range between 0 and 1. The integrated BPD, which consists of two germanium doped PIN photodiodes~\cite{hai_16_2013,Tait_2019}, is connected to a control circuit that only allows sensing and sourcing to occur at the same circuit location. This configuration results in an incorrect biasing voltage across the photodiodes which leads to larger noise in the output photocurrent compared to the off-chip BPD measurements. The signal fidelity of the on-chip BPD could be improved using a custom-designed circuit board that allows sensing and biasing from different locations within the circuit, similar to the design of the off-chip BPD.

We recorded two separate sets of 5000 photonic inner product measurements using the off-chip and on-chip BPD circuits by randomly varying the input vectors from the four laser sources (by modulating the signal amplitude) and the weight values of the MRRs (by tuning the resonance peaks using an applied current). The MRRs were initially calibrated to determine the mapping between the applied heating current and the corresponding optical weighting value. At each time step, the output photocurrent from the BPD was measured after all four channels were modulated with their corresponding inputs and weights.  This experiment is representative of typical dense matrix multiplication operations as all elements in the weight and input vectors were updated simultaneously. It accurately accounts for the device-to-device variability of the system, optical and electronic noise, crosstalk between neighboring MRRs, and propagation loss throughout the system. The measured output photocurrents for both circuits, scaled between $-1$ and 1, as functions of the expected multiplication output are shown in Fig.~\ref{fig:experimental}(a). The error in the experimental inner product operations of the off-chip photodetection circuit has a standard deviation of 0.098 (4.35 bits) and mean of 0.003, while the integrated BPD circuit has a larger standard deviation of 0.202 (effective resolution of 3.31 bits) and mean of 0.003. Further details concerning our experimental procedures are given in Supplement~1.

We then used Python simulations to train neural networks with the photonic inner product operation results from our two sets of measurements. The simulations add accurately scaled Gaussian noise, which represents the error in our experimental inner product measurements, to the output of each MAC operation in the matrix-vector multiplication for calculating the gradient~$\boldsymbol{\delta}^{(k)}$. The inference and weight update steps were performed using full-precision data operations as these can be executed using digital circuitry in our proposed architecture. Using a neural network of size $784\times800\times800\times10$, the simulation added Gaussian noise to the output of the matrix-vector multiplications between the fixed random matrix $\mathbf{B}^{(k)}$ $(800\times10)$ and error vector $\mathbf{e}$ (10$\times$1). Therefore, each matrix-vector multiplication performed 800 inner product operations between the error vector $\mathbf{e}$ and each row in the matrix $\mathbf{B}^{(k)}$. We trained this neural network on the MNIST dataset using ReLU activations in the two hidden layers and a softmax activation at the output. We used the cross-entropy loss function and the stochastic gradient descent optimization algorithm with a constant learning rate of 0.01, a momentum value of 0.9, and a mini-batch size of 64. Each simulation was run 10 times using different randomly initialized network parameters, and the mean results were recorded. The implementation of the Python simulation is further described in Supplement~1.

The validation accuracy during training using the experimental results, as well as without injecting noise, is shown in Fig.~\ref{fig:experimental}(b). The simulations using the off-chip and on-chip BPD circuits achieved test accuracies of $97.41 \pm 0.15\%$ and $96.33  \pm 0.16\%$, respectively, while a test accuracy of $98.10 \pm 0.13\%$ was achieved without injecting noise.
The test accuracy as a function of the effective resolution used in the matrix-vector multiplication operations is shown in Fig.~\ref{fig:experimental}(c). These simulation results demonstrate that the DFA algorithm is remarkably robust to noise and low-precision computations. Indeed, competitive performance using the DFA algorithm has even been shown using error values ternarized to $\{-1;0 ;1\}$ compared to full-precision training~\cite{launay_hardware_2020}. A similar technique of adding gradient noise when training very deep architectures with backpropagation has shown to avoid overfitting and lower the training loss by encouraging active exploration of the parameter space~\cite{neelakantan_adding_2015}. The remarkable robustness of neural networks to distortions against additive and multiplicative noise during inference and training~\cite{merolla_deep_2016} establishes analog photonic hardware as a promising platform for executing in situ training.

%Indeed, the DFA algorithm does not require high-precision or floating point numbers when calculating the gradient vectors; competitive performance has even been demonstrated using error values ternarized to $\{-1;0 ;1\}$ compared to full-precision training~\cite{launay_hardware_2020}.

%\red{Discuss the noise and explain why we see what we see (relatively good results with increased noise and low precision). We can cite a paper or two on “noise-aware training.... We must emphasize that hardware on-chip training is well suited for  “noise-aware training.“”}

\blue{ 

}

\section{Energy and speed analysis}

This section evaluates our proposed architecture's expected energy consumption and speed using high-performance electronic and optical components (in contrast, our current experimental system has an estimated energy consumption of ${\sim}\SI{2.0}{\micro\joule}$ per MAC operation as the thermally-tuned MRR weights have a slow tuning speed of $\SI{170}{\micro\second}$~\cite{Bogaerts_2012}).  Matrix-vector multiplications are the fundamental operation performed during neural network training, and  although they are computationally expensive to perform in digital electronics, they can be executed in a single operational cycle in photonics. The power efficiency and speed of the photonic architecture are dependent on the size of the photonic weight bank and the maximum operational rate $f_s$, which is limited by the electronic device on-chip with the lowest throughput \cite{Bangari_2020}. Defining an operation as either a multiplication or addition of two inputs, the maximum number of operations per second (OPS) computed by the photonic architecture scales linearly with the size of the weight bank ($M\times N$): 
\begin{equation}
\label{eq:OPS}
    \mathrm{OPS} = 2f_s MN.
\end{equation}

The lower bound on the required optical power per laser supplied to the photonic weight bank to overcome both the capacitance of the photodetectors with $N_b$ bits of fixed precision and the shot noise is given by
\begin{equation} \label{eq:E_op_MAC}
P_{\mathrm{laser}} \geq M \frac{\hbar 
\omega}{\eta} \max \left(2^{2 N_{b}+1}, \frac{C V_{d}}{e}\right),
\end{equation}
where $\hbar\omega$ is the photon energy, $\eta$ is the combined quantum efficiency of the laser, photodetector, and optical system loss through the waveguide, $C$ and $V_d$ are the capacitance and driving voltages of the photodetectors, and $e$ is the elementary charge~\cite{Nahmias_2020}. 

\begin{figure}[t]
    \centering
    \includegraphics{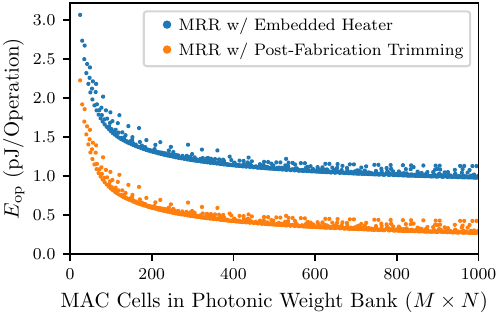}
    \caption{The expected optimal energy per operation $E_\mathrm{op}$ as a function of the number of MAC cells in the photonic weight bank using both embedded heaters for MRR resonance locking and post-fabrication trimming ($M, N \ge 5$). The total energy consumption is dependent on both the number of MAC cells and the dimensions of the photonic weight bank, and only the lowest value of $E_\mathrm{op}$ as a function of the total number of MAC cells (i.e., the ideal photonic weight bank dimensions) is shown. The calculated values assume an operating speed of 10 GHz and 6 bits of precision.}
    \label{fig:E_op}
\end{figure}

During training, the tunable MRRs in the architecture are inscribed with a different value at each operational cycle. The values in the error vector $\mathbf{e}$, which can be stored in static random-access memory (SRAM), are fetched and converted into analog voltage signals for tuning the designated MRRs using digital-to-analog converters (DACs). Unlike the dynamic $\mathbf{e}$ vectors that change at each operational cycle, the inscribed weighting values of the MRRs in the photonic weight bank cycle through a set of known, unchanging values from the matrices $\mathbf{B}^{(k)}$ for each hidden layer $k$. The control system for these MRRs can be made highly efficient by storing the set of weight values in an analog memory architecture, such as using resistive random-access memory (RRAM) or optical memories based on phase change materials \cite{Zahoor_2020, rios_integrated_2015}, where the energy consumption required to access and switch between weights is negligible compared to the energy cost of modulating the input optical signals with the error vector. 

The power required to control the transmission state of a single MRR, $P_{\mathrm{MRR}}$, includes two components: the power required to tune the MRR on and off-resonance to achieve the desired weighting value and the power required to lock the MRR weight onto resonance. The first component is negligible when using carrier depletion for high-speed tuning ($\sim$$\SI{120}{\micro W}$)~\cite{tait_quantifying_2022, timurdogan_ultralow_2014}. The second component is necessary due to fabrication nonidealities which cause a resonance shift; this shift can be greater than the tuning range allowed via carrier depletion, so embedded N-doped heaters are often used for thermal locking~\cite{Peng_Nahmias_2018, al-qadasi_scaling_2022}. However, heaters require $\sim$14~mW, which can result in significant overall energy consumption~\cite{tait_quantifying_2022, Jayatilleka_2015}. Alternatively, post-fabrication non-volatile phase trimming techniques can be used to correct for fabrication variations by modifying the refractive index of the waveguide or cladding, eliminating the need for thermal locking~\cite{jayatilleka_post-fabrication_2021}. 

The estimated wall-plug power required by the DFA architecture with a photonic weight bank of size $M\times N$ is given by
\begin{equation}
\label{eq:P_total}
P_\mathrm{total} = NP_{\mathrm{laser}}
+N(M+1)P_{\mathrm{MRR}}
+NP_{\mathrm{DAC}}
+M( P_{\mathrm{TIA}}
+P_{\mathrm{ADC}}).
\end{equation}
The first term is the minimum necessary optical power, as shown in Eq.~\eqref{eq:E_op_MAC}. The second term is the heating power required to tune the MRRs in the architecture as previously discussed.  The third term accounts for the power required to convert the $\mathbf{e}$ vector values into analog control signals using DACs ($P_{\mathrm{DAC}}$), and we assume that the energy required to fetch the data from the SRAM is negligible compared to the other components in the architecture \cite{low_power_sram}. The final term corresponds to the power from the TIAs ($P_{\mathrm{TIA}}$) and ADCs ($P_{\mathrm{ADC}}$).

Using Eqs.~\eqref{eq:OPS} and \eqref{eq:P_total}, the expected optimal energy consumption per operation $E_\mathrm{op}$, defined by $P_\mathrm{total}/\mathrm{OPS}$, as a function of the number of MAC cells in the photonic weight bank is shown in Fig.~\ref{fig:E_op}. The energy consumption using both embedded heaters for MRR locking (14.12 mW per MRR), as well as post-fabrication trimming of the MRRs ($\SI{120}{\micro W}$ per MRR), is shown. To achieve an operational rate in the GHz range with low-power consumption, our calculations assume the use of efficient, high-speed MRRs with small footprints that are tuned using carrier depletion in an embedded reverse-biased PN-junction \cite{Dong_2010, sun_128_2019}. The power and performance speed of the photonic architecture were calculated assuming the use of the ReLU activation function and the following energy consumptions for the active electronic components:  
180 mW per DAC (12 bit, 10 GS/s, Alphacore D12B10G), 
13 mW per ADC (6 bit, 12 GS/s, Alphacore A6B12G), 
and 2.4 pJ/bit per TIA (20 GS/s) \cite{TIA_properties}. 
The throughput of the DAC limited the maximum operational rate $f_s$ to 10 GHz. We assumed an efficiency $\eta$ of 0.2 for the laser source, photodetector, and optical system loss through the waveguide. The calculations assume the use of an optical signal with a wavelength of 1550~nm and a high-performance photodetector with a capacitance of 2.4~fF and a driving voltage of 1~V \cite{Feldmann_2021}. Using a  photonic weight bank of size $50\times20$, we can achieve speeds of 20 teraoperations per second (TOPS) and an energy consumption $E_\mathrm{op}$ of 1.0 pJ per operation using MRRs with thermal heaters and 0.28 pJ per operation using post-fabrication trimming of the MRRs. The estimated compute density, defined by the OPS divided by the chip area, is 5.78 TOPS/mm$^2$. We assume a photonic MAC cell of size $\SI{47.4}{\micro\metre} \times \SI{73.0}{\micro\metre}$, which accounts for the overhead waveguide and electronic routing, bonding pads for MRR control, and sufficient spacing between MRRs to eliminate crosstalk (see Fig.~\ref{fig:mrrs}(a)). 

 \section{Discussion}  

 The recent proliferation of analog hardware-based neural networks has resulted in several proposed photonic architectures that have the potential to outperform state-of-the-art digital systems for inference computations~\cite{bernstein_single-shot_2022, Feldmann_2021, xu_11_2021, ashtiani_-chip_2022}.  However, these photonic architectures are primarily trained offline using algorithms implemented on conventional digital platforms that attempt to simulate the analog hardware, resulting in longer training times due to the significant computational overhead. Additionally, the simulations may not accurately model all the nonidealities and noise sources in the photonic system, resulting in lower on-chip inference accuracy than expected by the simulated models.  Motivated by these constraints, we have introduced a silicon photonic architecture that executes the fundamental operations of the DFA algorithm in the analog domain, including matrix-vector multiplication and the Hadamard product, to calculate the gradient vector for each neural network layer. Our proposed approach enables in situ training directly on a PIC, which can inherently account for nonidealities in the analog hardware while taking advantage of the high-bandwidth and low-energy consumption offered by photonics. Additionally, in situ training with photonic hardware can enable training with data signals originally generated in the optical domain, eliminating the need for optical-to-electronic conversions. 
 
The photonic architecture is highly scalable for training neural networks of variable sizes due to its compatibility with GeMM compilers, which subdivide and process the on-chip matrix-vector operations based on the integrated photonic weight bank dimensions. Although analog photonic circuits inherently contain noise and cannot match the precision offered by digital electronics, neural networks exhibit remarkable robustness to distortions against additive and multiplicative noise during inference and training~\cite{merolla_deep_2016}. Indeed, as demonstrated by our simulation results, the DFA training algorithm performs well even with added noise during the calculation of the gradient vector (0.69\% and 1.77\% drop in accuracy on the MNIST dataset using inner product operation measurements from the off-chip and on-chip BPD circuits, respectively). The DFA algorithm is particularly well suited for implementations with analog hardware as the gradient vector is calculated by propagating the error through fixed random feedback connections directly from the output layer to each hidden layer, which is advantageous as noise does not accumulate between layers~\cite{nokland2016direct}. This is unlike the backpropagation algorithm, where the error is back-propagated layer by layer from the output layer to the hidden layers.

Integrated photonics is a promising platform for implementing machine learning algorithms as the high-throughput and low-latency offered by PICs could enable the implementation of ultra-fast and highly efficient analog neural networks. Our proposed architecture can perform the gradient vector computation in a single time step by incorporating WDM techniques to process multi-channel analog signals along the same waveguide bus. Using a weight bank of size $50\times20$ with high-performance photonic and electronic components, the architecture is expected to perform up to 20 TOPS (limited by the signal modulation and detection speeds) while consuming less than 1 pJ per MAC operation and can be scaled up in terms of wavelength parallelization, number of vector multipliers, and signal baudrate. The expected improvements in training time and energy efficiency offered by our proposed architecture could enable the development of innovative neural network applications that cannot operate on current generation hardware. Future work includes demonstrating a complete integrated system with a dedicated CMOS processor capable of operating at high-speeds for neural network training without requiring any data processing off-chip.

\section*{Acknowledgements}
This research is supported by the Natural Sciences and Engineering Research Council of Canada (NSERC) and the Vector Scholarship in Artificial Intelligence, provided through the Vector Institute. V.J.S. is supported by the AFOSR PECASE Award under contract number (FA9550-20-1-0193).  

\section*{Disclosures}
The authors declare no competing interests.

\section*{Data availability}
The data that support the findings of this study are available from the corresponding author upon reasonable request.

\section*{Supplemental document}
See Supplement~1 for supporting content.

\end{document}